\documentclass[10pt]{article}
\usepackage[utf8]{inputenc}
\usepackage[T1]{fontenc}
\usepackage{amsmath}
\usepackage{amsfonts}
\usepackage{amssymb}
\usepackage[version=4]{mhchem}
\usepackage{stmaryrd}
\usepackage{graphicx}
\usepackage[export]{adjustbox}
\graphicspath{ {./images/} }
\usepackage{hyperref}
\hypersetup{colorlinks=true, linkcolor=blue, filecolor=magenta, urlcolor=cyan,}
\usepackage{algorithm,algorithmic}

\title{Boosting Local Spectro-Temporal Features for Speech Analysis\footnote{Master's project, University of Toronto. Advisor: $\text{Allan}_{+}$ D. Jepson.}}

\author{Michael Guerzhoy}

\date{June 2010}

\begin{document}
\maketitle

\begin{abstract}
We introduce the problem of phone classification in the context of speech recognition and explore several sets of local spectro-temporal features that can be used for phone classification. In particular, we present some preliminary results for phone classification using two sets of features that are commonly used for object detection: Haar features and SVM-classified Histograms of Gradients (HoG).
\end{abstract}

\section{Introduction}
Speech recognition can proceed, roughly, as follows. The speech data is broken up into short time segments (shorter than the duration of almost all phones) of equal durations; features are extracted from each segment; the segments are classified based on the extracted features (the output of the classifier may be confidence-rated); finally, a graphical model is used to obtain phone labels for each segment. The graphical model can be trivial, or it can segment the phones explicitly. Typically, Hidden Markov Models (HMM) are used.

In this report, we concern ourselves with the phone classification part of the task.

While in general, accurate segmentation of the speech data into phones is not available, our experiments are performed on hand-segmented phone data from the TIMIT dataset~\cite{timit}. The availability of the segmentation makes the phone classification task easier, but also introduces the problem of dealing with sound samples of variable length, since phones can have different durations.

This report is organized as follows.

We first discuss spectrograms, the standard initial transformation of sound signals.

We then discuss Mel-frequency cepstral coefficients (MFCC), the standard feature set for phone classification, and how these features are classified.

We then discuss local spectro-temporal features and compare them to MFCC.

We then describe our approach to phone classification, which consists of boosting binary stump classifiers of local spectro-temporal features with AdaBoost to classify a segment as belonging to one of two possible phones, and then using $N(N-1) / 2$ such classifiers, where $\mathrm{N}$ is the number of possible phones, to obtain a label for the segment.

We discuss two feature sets used in object detection: Haar features and Histograms of Gradients (HoG), and how they can be used in our approach.

We describe some experiments performed using that approach.

\section{Spectrograms}
A digital sound signal $x(t)$ is a time series, air pressure recorded at regular intervals of time.

The first step in obtaining features for speech analysis is obtaining the spectrogram of the signal. The spectrogram is the squared magnitude of the coefficients obtained by applying a short-time Fourier transform (STFT) to the signal~\cite{huang2001}.

The computation is performed by computing the Discrete Fourier Transform of frames in the signal (we follow the presentation in Huang et al.~\cite{huang2001}). Frame $m$ of the signal is defined as

$$
x_{m}[n]=x[n] w\left[c_{m}-n\right], \mathrm{c}_{\mathrm{m}}=\frac{N}{2}+\mathrm{i} \cdot \mathrm{m},
$$

where $w[n]$ is nonzero only for $|n|<N / 2$. $N$ is referred to as the length of the frame and $i$ is a positive integer referred to as the increment. Many choices are possible for the form of $w$. We use the Hamming window:

$$
w[n]=\left\{\begin{array}{c}
0.54-0.46 \cos \left(\frac{2 \pi n}{N-1}\right), 0 \leq n<N \\
0, \text { otherwise }
\end{array}\right.
$$

We compute $X_{m}$, the Discrete Fourier Transform of \\ $\left[x_{m}\left[c_{m}-\frac{N}{2}\right], \ldots, x_{m}\left[c_{m}+\frac{N}{2}-1\right]\right.$. The m-th column of the spectrogram is then the magnitudes of the coefficients of $X_{m}$. The length of the frame $\mathrm{N}$ must be chosen when computing the spectrogram. There is a trade-off between good resolution in time and good resolution in frequency when choosing this length.

\begin{figure}
\begin{center}
\includegraphics[max width=\textwidth]{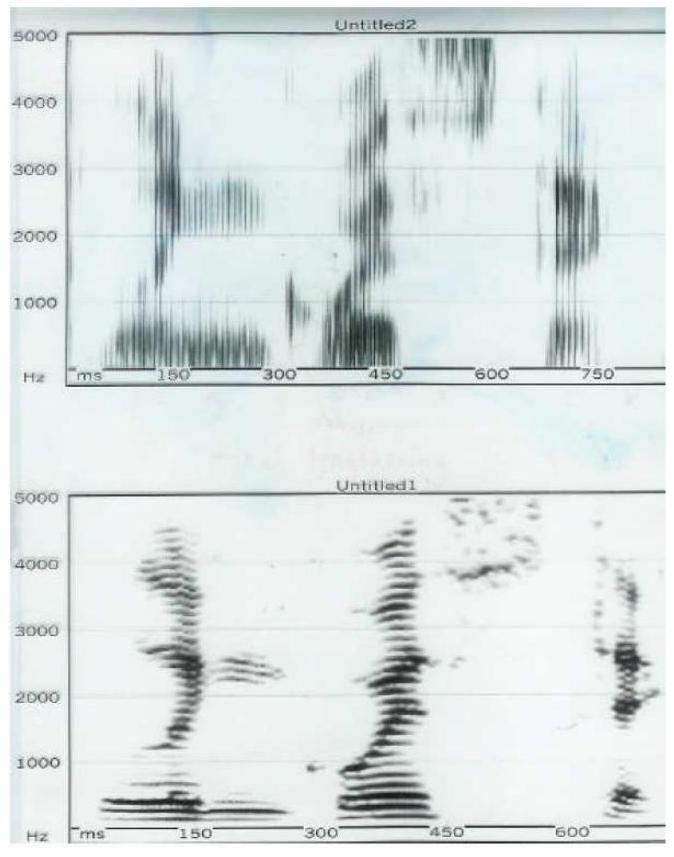}
\end{center}

\caption{\label{fig1} Wideband spectrogram (top) and narrowband spectrogram (bottom). Source: \url{http://www.linguistics.ucsb.edu/faculty/gordon/acousticpdf/widebandlinguistics.pdf}}
\end{figure}

Spectrograms computed using a short (relative to the carrier wavelength of the signal) time window are wide-band spectrograms (see the top image in Figure~\ref{fig1}). When analyzing a speech signal, vertical lines (``striations") typically appear in wide-band spectrograms. They appear because the sampling window is shorter than twice the fundamental frequency of the source signal. More importantly, changes over time are more clearly visible in the wide-band spectrogram.

Using a longer time window results in a narrow-band spectrogram (see the bottom image in Figure~\ref{fig1}). Typically, the harmonics of the source signal would be visible in the narrow-band spectrogram but not in the wide-band spectrogram.

\section{MFCC}
\subsection{MFCC coefficients \label{mfcccoeffs}}
\begin{figure}
\begin{center}
\includegraphics[max width=\textwidth, center]{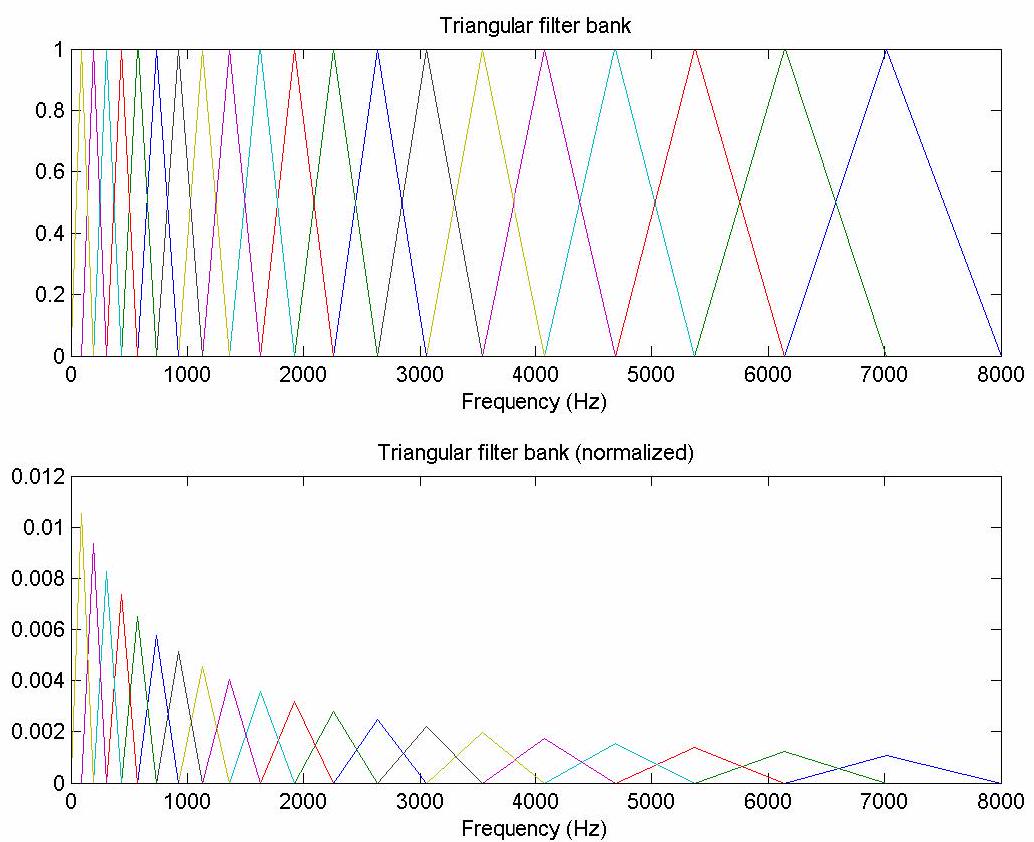}  
\end{center}
\caption{\label{tribank}Triangular filter banks. Source: \url{http://neural.cs.nthu.edu.tw/jang/books/audioSignalProcessing/example/speechFeature/output/showTriFilterBank01.png}}
\end{figure}

Mel-frequency cepstral coefficients (MFCC) coefficients are defined for each time $t$ and are derived from the spectrogram~\cite{huang2001}. A triangular filter bank across frequencies, like the one pictured in Fig.~\ref{tribank}, is applied to a time-slice of the spectrogram and the log of the outputs of the filter bank is computed. The MFCC coefficients are then the first $~ 16$ coefficients of the DCT of the logs of outputs of the filter banks (higher-order coefficients can be obtained but do not improve performance~\cite{huang2001}).

Applying the triangular filter bank effectively makes the sampling of the spectrogram finer for lower frequencies and coarser for higher frequencies. In other words, the frequency axis is logarithmically rescaled.

The transformations of the spectrogram before the DCT is performed improve performance because the coefficients for higher frequencies are always less reliable (so it makes sense to smooth the high-frequency coefficients more, since averaging many coefficients makes the result more reliable), and because the range of the coefficients is very large (so it makes sense to take the log before applying the DCT so that outlier coefficients in the spectrogram don't influence the DCT coefficients too much).

Logarithmically rescaling the frequency axis and computing the logarithm of the spectrogram can be justified by the fact that Weber's law holds for both the frequency and intensity of sound~\cite{moore2003}: the smallest change detectable by a person in frequency (resp. intensity) of sound is proportional to the initial frequency (resp. intensity).

\subsection{MFCC-based dynamic features}
Since the spectrograms for some phones change over time, it is desirable to add features that reflect the change in the spectrogram over time. To that end, the delta MFCC coefficients are used. Essentially, the delta coefficients are finite-difference approximations of the first and second time-derivatives of the MFCC features. The delta MFCC significantly improve classification performance compared to the performance when using only MFCC features.

\subsection{MFCC-based phone classification}
A standard approach to phone classification is to fit a mixture of Gaussians in the MFCC feature space to each phone, and then to obtain the probability for each phone by extracting the features and computing the output of the mixture of Gaussians on the features. A discriminative approach is also possible. For example, Clarkson and Moreno~\cite{clarkson1999} use all-vs.-all (see Section~\ref{allVall} for details) classification based on SVM binary classifiers to obtain state-of-the-art results.

\section{Spectro-Temporal Patch Features for Speech Analysis}
\subsection{Motivation}
MFCC features work well in practice, but there are reasons to think that they are not optimal for phone classification.

First, MFCC features are generally low-dimensional. (More than the first 15 MFCC coefficients could be used, but in practice this does not improve performance~\cite{huang2001}). In principle, it is possible that information is lost by using the low-dimensional MFCC feature vector, and using features with more dimensions could improve performance.

Second, MFCC features are global (in the frequency but not the time dimension). The entire time-slice of the spectrogram is used when computing MFCC coefficients. This means that noise in a patch in the spectrogram could potentially influence the entire MFCC feature vector (note, however, that our experiments are performed on speech with no significant noise).

Using the outputs of filters with local support within the spectrogram addresses both of those issues. The filters can be applied at all the points in the spectrogram segments, so we obtain a very high-dimensional feature vector for each spectrogram segment. Since the filters have local support, noise and outliers in the spectrogram can only influence a limited number of coefficients in the feature vector. This can potentially improve classification performance.

Edge features are promising for use in classification. Vowels as well as some consonants are characterized by high energy bands in the spectrogram (the formants), so features which indicate local horizontal edges should be useful. Plosives are characterized by vertical (i.e., temporal) edges in the spectrogram, so local vertical edges detectors can be useful. Diagonal edges are useful for classifying, e.g., the sound of some approximants: for example, the second format of /w/ is rising with time.

\subsection{Previous work on phone classification using spectro-temporal patch features \label{PrevWorkClassSpectroTemp}}
\begin{figure}
\begin{center}
\includegraphics[max width=\textwidth]{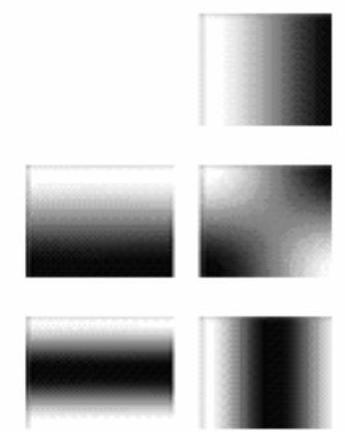}
\end{center}
\caption{\label{2ddct} The first 6 basis functions of 2D-DCT. Source: Bouvrie et al.~\cite{bouvrie2008}}

\end{figure}

Bouvrie et al.~\cite{bouvrie2008} extract the first 6 2D-DCT coefficients of small patches of the spectrogram arranged on a grid (Figure~\ref{2ddct})

Since the spectrograms are of different lengths, Bouvrie et al. do the following to obtain a feature vector of constant length. The coefficients are extracted from patches lying on a grid on the spectrogram. The spectrogram is then divided into 5 segments of length proportionate to the length of the spectrogram along the time axis, and the 2D-DCT coefficients of patches of the same size, the same frequency coordinate, and lying within the same segment are all averaged together. This way, regardless of the length of the spectrogram, we obtain a constant number of coefficients from each segment, and, concatenating the coefficients obtained from all five segments, we obtain a feature vector of constant size (about 500).

Bouvrie et al. use the length of the sample as a feature as well. The length is concatenated to the feature vector described above.

For $N=49$ phones, $N^(N-1) / 2$ classifiers are trained for each pair of phones. When classifying a sample, Bouvrie et al. use all-vs.-all classification. All $N(N-1) / 2$ classifiers are run on the sample, and the vote count for the ``winning'' phone is incremented by 1 . The output of the classifier is the phone with the highest vote count. Bouvrie et al. claim that all-vs.-all is the optimal method for multi-class classification, even though more sophisticated methods are available.

Bouvrie et al.'s classifier produces a 25\% error rate when using linear classifiers when classifying pairs of phones and a $21 \%$ error rate when using tuned nonlinear SVM classifiers on the TIMIT dataset.

Amit et al.~\cite{amit2005} extract binary edge features from the spectrogram (essentially; the edge features are replicated locally), and then classify phones using maximum likelihood after estimating the likelihood of each phone using a trained Naïve Bayes classifier where the features are the presence or absence of an edge point at a specified coordinate. Amit et al. pad the spectrograms with zeros at the end if they are shorter than $200\mathrm{~ms}$ in length and truncate them to $200\mathrm{~ms}$ if they are longer. By doing this, they implicitly provide the classifier with information about the length of the sample. The major advantage of this approach is that it requires a very small training set, since no correlations between features are taken into account when training. On the other hand, the performance suffers for that exact reason.

Amit et al. obtain 41.5\% classification error rate on the TIMIT dataset.

\section{Boosting Local Spectro-Temporal Features for Phone Classification \label{method}}
\subsection{Introduction}
We propose to build a system for phone classification similar to the one used by Viola and Jones~\cite{viola2002} for face detection. Given a spectrogram of a given height and variable length (in pixels), we aim to determine which phone it corresponds to. As with the first steps to compute MFCC coefficients, the spectrograms are rescaled to mel-scale in our experiments, the logarithm of the spectrogram is taken, and the values are clipped and then rescaled to be within $[0,1]$.

We report experiments using two sets of image features: Haar features~\cite{viola2002} and Histograms of Gradients classified with linear support vector machines (SVM)~\cite{zhu2006}.

We first discuss AdaBoost. We then discuss the two feature sets in our experiments: Haar features and SVM-classified HoGs. We then discuss several strategies for dealing with the fact that the phones are of variable length. We then discuss how to do multi-class classification from the response of the trained binary classifiers for pairs of phones. Finally, we present some experimental results.

Our samples are of variable length. In order for our classification methods to work, we need to be able to have a one-to-one correspondence between features extracted from different samples. In what follows, we will occasionally gloss over this issue by tacitly assuming that all the samples (and their spectrograms) are of the same length (in which case features can be characterized by their coordinates on the spectrogram, their shape, and their scale). We discuss how we handle the issue of the variable length of samples in Section ~\ref{varlength57}.

In the discussion that follows, the task under discussion is binary classification (specifically, classifying a sample as either of two phones). The full phone classification task, however, requires multi-class classification. In Section~\ref{multiclassclass}, we discuss how to use binary classifiers for the full phone classification task.

\subsection{Boosting}
A boosting algorithm is an algorithm for learning a ``strong'' classifier by combining many ``weak classifiers,'' given training data with target outputs.

The basic idea behind boosting is to learn many different weak classifiers, select some of them iteratively, and combine them into a strong classifier by computing a weighted sum of their outputs. If the weak classifiers are more or less independent (i.e., not all of them are wrong on the same samples), we can combine them into a strong classifier.

The weak classifiers are functions of the features. They have a specific form. For instance, if the samples in the input data are $n$-dimensional vectors, we might specify that the weak classifiers are of the form

$f_{i}(x)=\left\{\begin{array}{l}1, \text { if } x[i]>t_{i} \\ -1, \text { otherwise }\end{array}\right.$

for some coordinate $\mathrm{i}$ and threshold $t_{i}$

We use weak classifiers of the above form in our experiments.

\subsection{Discrete AdaBoost and Gentle AdaBoost}
The originally-proposed AdaBoost algorithm, Discrete AdaBoost, combines weak classifiers as follows.

Each sample in the training set is initially assigned equal weight. At each iteration, a pool of available weak classifiers is created by training them (e.g., setting the thresholds $t_{i}$ in the example above to minimize the weighted error over the samples). Out of the pool of available weak classifiers, we select the one which minimizes the weighted error. The weighted error is the sum of the classification errors (the error is 1 if the prediction is incorrect and 0 if it is correct) for the individual samples in the training set times the weights assigned to them. Equivalently, we can minimize the expected error $E_{w}$ (see Algorithm~\ref{DiscAda}), which is just the same sum divided by the number of samples. After the weighted error-minimizing weak classifier is selected, the samples in the training set are re-weighted such that the samples on which the currently-selected weak classifier is wrong are weighted more than at the previous iteration, and the samples on which it is correct are weighted less than at the previous iteration.

As a result of the re-weighting, the samples which are currently classified incorrectly will be given more weight when selecting the next weak classifiers.

Given training data $x_{1}, x_{2}, \ldots, x_{N}$ and target outputs $y_{1}, y_{2}, \ldots, y_{N}$ with $\mathrm{y}$ in $\{-1$, $1\}$, the Discrete AdaBoost algorithm used in the experiments below is as follows~\cite{freund1996experiments}.

\begin{algorithm}[h]
\caption{Discrete AdaBoost(Freund \& Schapire~\cite{freund1996experiments}}
\begin{algorithmic}  
  
    \STATE Start with weights $w_{i}=1 / N, i=1, \ldots, N$.
    
     \FOR{$m=1,2, \ldots, M$}

        \STATE Fit the classifier $f_{m}(x)$ using weights $w_{i}$ on the training data.
        \STATE Compute $e_{m}=E_{w}\left[1_{\left(y \neq f_{m}(x)\right)}\right], c_{m}=\log \left(\left(1-e_{m}\right) / e_{m}\right)$.
        \STATE Set $w_{i} \leftarrow w_{i} \exp \left[c_{m} \cdot 1_{\left(y_{i} \neq f_{m}\left(x_{i}\right)\right)}\right], i=1,2, \ldots N$, and renormalize so that $\sum_{i} w_{i}=1$.
  \ENDFOR{}
  \STATE Output the classifier $\operatorname{sign}\left[\sum_{m=1}^{M} c_{m} f_{m}(x)\right]$
  
\end{algorithmic}

 \caption{\label{DiscAda} $E_{w}$ represents expectation over the training data with weights $w=\left(w_{1}, w_{2}, \ldots w_{n}\right)$. At each iteration AdaBoost increases the weights of the observations misclassified by $f_{m}(x)$ by a factor that depends on the weighted training error. }

\end{algorithm}

Sometimes we cannot find a classifier $f_{m}(x)$ such that $e_{m}<0.5$. In that case the algorithm terminates.

Given training data  $x_{1}, x_{2}, \ldots, x_{N}$ and target outputs 
$y_{1}, y_{2}, \ldots, y_{N}$ with $y \in \{-1$, $1\}$, 
the Gentle AdaBoost algorithm used in the experiments below is as follows.

\begin{algorithm}[h]

\begin{algorithmic}
  \STATE Start with weights $w_{i}=1 / N, i=1,2, \ldots, N, F(x)=0$
  \FOR{$m=1,2, \ldots, M$}
    \STATE Estimate $f_{m}(x)$ by weighted least-squares fitting of $y$ to $x$.
    \STATE Update $F(x) \leftarrow F(x)+f_{m}(x)$
    \STATE Update $w_{i} \leftarrow w_{i} e^{-y_{i} f_{m}\left(x_{i}\right)}$ and renormalize.
  \ENDFOR{}
  \STATE Output the classifier $\operatorname{sign}[F(x)]=\operatorname{sign}\left[\sum_{m=1}^{M} f_{m}(x)\right]$
\end{algorithmic}
\caption{Gentle AdaBoost (from Friedman et al.~\cite{friedman1998})}
asdf
\end{algorithm}

Note that the difference between the Discrete and Gentle variants of AdaBoost is merely in the computation of the weights and the criteria for selecting the weak classifiers. In both cases, we consider all possible features.

We use the Gentle AdaBoost method in our experiments.

\subsection{AdaBoost as a feature selection procedure}
When the weak classifier pool is extremely large (as is the case in our system), only a small proportion of the weak classifiers is selected by AdaBoost. If the weak classifiers are functions of only a few coordinates of the input sample, we can view the selection of weak classifiers by AdaBoost as the selection of informative features.

\subsection{Haar features}
Boosted Haar features were used by Viola et al.~\cite{viola2002} for face detection. At a given location in the spectrogram, and at a given scale (i.e., size of the filter), one of the following filters is applied to obtain a features (images taken from~\cite{lienhart2002}):

\begin{figure}[h]
\includegraphics[max width=\textwidth, center]{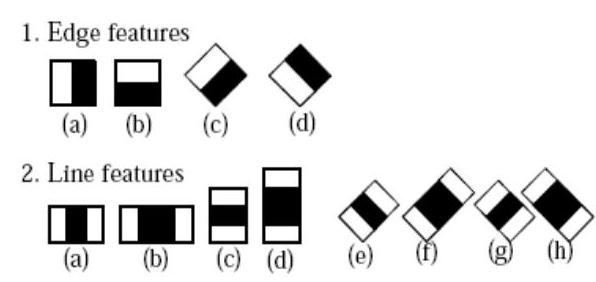}

\begin{enumerate}
  \setcounter{enumi}{2}
  \item Center-surround features
\end{enumerate}

\begin{center}
\includegraphics[max width=\textwidth]{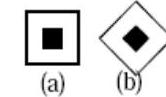}
\end{center}

\begin{enumerate}
  \setcounter{enumi}{3}
  \item Special diagonal line feature used in $[3,4,5]$
\end{enumerate}

\caption{Feature prototypes of simple Haar-like and center-surround features. Black areas have negative and white areas positive weights. From Lienhart et al.~\cite{lienhart2002}}

\end{figure}

The weights are such that when the filters are applied to a constant image, the output is 0.

To obtain the pool of the features, we systematically apply filters of all types at all locations and at many scales.

\subsection{SVM-classified Histograms of Gradients}
SVM-classified Histogram of Gradients features (HoG) were used by Zhu et al.~\cite{zhu2006} for pedestrian detection.

HoG features are computed as follows.

On a patch with the upper-left corner at coordinates $(i, j)$ in the image and of width $w$ and height $h$, the HoG feature is computed as follows. The Histogram of Gradients is a 9-bin histogram, with the bins representing a range of edge directions $\left\{\left[0 \ldots \frac{1}{9} 2 \pi\right),\left[\frac{1}{9} 2 \pi \ldots \frac{2}{9} 2 \pi\right), \ldots,\left[\frac{8}{9} 2 \pi \ldots 2 \pi\right)\right\}$. First, a patch of a given scale at a given location in the spectrogram is extracted from the spectrogram $S$ (note that the patch is not rescaled). At each location $(x, y)$ in the extracted patch, we compute $\mathrm{dx}=$ $S(x+1)-S(x-1)$ and $d y=S(y+1)-S(y-1)$. We then obtain the direction of the vector $(dx, dy)$ and the magnitude of the vector $\sqrt{d x^{2}+d y^{2}}$. The magnitude is divided by the distance from the centre of the image patch, and the result is added to the appropriate bin for the direction of (dx, dy). Finally, the histogram is normalized by dividing the sum in each bin by the largest sum in the histogram.

We extract features at all locations and for widths and heights $(t, t),(2 t, t)$, and (t, 2t) for many values $t$ in $\{2,4,6, \ldots\}$ such that the patches are all smaller than the image.

Note that Zhu et al.~\cite{zhu2006} use more complicated features: they group the features we use into $2 \times 2$ blocks, and concatenate the histogram in each of the 4 cells in the block before normalizing the histograms using the maximum sum over all 4 cells. We do not do this since this leads to a larger minimum size for features and did not seem to help with classification.

After the Histogram of Gradients is computed, it is fed into the SVM classifier that corresponds to it, and the SVM-classified HoG feature is the raw output of the SVM classifier.

There are various ways to train the SVM classifiers. We found that simply training a linear SVM classifier on the samples (with one phone corresponding to the output -1 and the other corresponding to the output 1) produces the same classification results as more complex variants (e.g., training nonlinear SVMs, using the weights obtained during the boosting process, etc.). We use this simple variant in the experiments.

\subsection{Variable-length samples \label{varlength57}}
In order to learn a classifier, we need to have a one-to-one correspondence between features extracted from different samples.

See Section~\ref{PrevWorkClassSpectroTemp} for a discussion of the ways this issue was handled in previous work. We explore four options for handling the issue.

\begin{enumerate}
  \item If the centre of the sample is $t$, the sample is considered to be $[t-dt, t+dt]$ for constant $dt$.

  \item The spectrogram of the sample is computed, and then warped to a constant size.

  \item The image to be treated as the spectrogram is actually $n=3$ stacked frames. Each frame corresponds to a range of time lengths (specifically, we use $0~ms$ to $75~ms$, $75~ms$ to $150~ms$, $150~ms$ and up). The spectrogram of the sample is computed and warped to be the size of one frame in whose range its length lies, and put in the frame. The two other frames are set to $0$ . 
  \item A standard length of a sample, T0, is specified. If the length of the sample T1 is larger than the minimum length, the feature corresponding to the one at coordinate (f, t) on the standard length of samples is the average (or the maximum) of the same type/size features computed at $(f, \text{floor} (t * T 1 / T 0)- 1)....(f, \text{ceil}(t*T1/T0))$

\end{enumerate}

\subsection{Multi-class classification \label{multiclassclass}}
Multi-class classification is the task of classifying a sample as belonging to one of $N$ classes (in our case, one of $N$ phones).

There are two basic alternatives to performing multi-class classification, one-vs.-all and all-vs.-all classification.

\subsubsection{One-vs.-all classification}
In one-vs.-all classification, $N$ classifiers are trained, one for each phone. For phone $p$, the positive samples are samples of the phone $p$ in the training set, and the negative samples are samples that do not contain the phone $p$ in the training set. On a given sample, the classifier associated with the phone $p$ outputs a real number that is larger if the classifier is more confident that the sample is p. The output of the one-vs.-all classifier is then the phone $f$ such that the output of the classifier that is associated with $f$ is the largest.

We do not report the full results for one-vs.-all classification. However, when we trained one-vs.-all classifiers, we have obtained classification results inferior to the ones we report for all-vs.-all classification.

\subsubsection{All-vs.-all classification \label{allVall}}
In all-vs.-all classification, $N(N-1) / 2$ classifiers are trained using the training set, one for each pair of phones. On a given test sample, each of the $\mathrm{N}(\mathrm{N}-1) / 2$ classifiers is run, and we assign a vote to the phone that's output by the classifier. The phone that gets the most votes is the output of the multi-class classifier.

\subsubsection{Better all-vs.-all classification \label{betterallVall}}
Most of the votes in all-vs.-all classification are meaningless: for example, if the sample is really /w/, it probably doesn't matter what the output of the classifier that classifies /s/ and /z/ is on this sample. This can, and in many cases does, cause the wrong phone to get the majority of votes.

A hierarchical voting procedure can ameliorate this problem. The first stage of this procedure reduces the set of phone candidates from $N$ to $N - 1$. The stage proceeds in $N -  N1$ iterations. At each iteration, the phone that got the least amount of votes is eliminated from consideration. We continue until $N1$ phones are left, and then run all-vs.all classification on those $N1$ phones.

\subsection{Discussion}
There are two justifications for using large pools of features.

First, the formants are slightly different for different speakers (e.g., they shift up and down depending on the speaker.) Therefore, if features at a fine enough scale are required, we expect that many copies of the same feature at neighbouring locations will be selected.

By selecting features from a large pool of features instead of using $\sim 500$ pre-selected features as in Bouvrie et al.'s work, we can gain an advantage accuracy (because better features are selected) or speed at test time (because unneeded features are not computed).

\section{Experiments}
We present some preliminary experimental results for the techniques proposed in Section~\ref{method}. For some of the experiments, Intel's OpenCV~\cite{bradski2000opencv} software, and in particular a modified version of the \texttt{haartraining} module of OpenCV, were used.

Below, we discuss some experiments using the TIMIT dataset~\cite{timit}. We use the standard train and test sets, and do not distinguish between the different speakers. We use 48 phones, and don't count confusions between $\{$ sil, cl, vcl, epi $\},\{$ el, 1$\},\{$ en, n , $\{$ sh, zh $\},\{$ ao, aa $\},\{\mathrm{ih}, \mathrm{ix}\},\{\mathrm{ah}, \mathrm{ax}\}$, as discussed in Lee et al.~\cite{lee1989}. The TIMIT data is sampled at $16000~Hz$. We present the full results for the full phone classification task for classification using Haar features on spectrograms that are warped to size $14 \times 15$ (i.e., $14$  log-frequency units times $15$ time units).

We also present some preliminary results for other configurations. We focus on several pairs of phones: t/d, s/sh, aa/ah, m/n, which are the hardest to distinguish since they share many phonetic features. For most pairs of phones, the classification error is always essentially 0.

\subsection{The full phone classification task (Haar features, warped spectrograms)}
We use the TIMIT segmentation to obtain the individual phones, obtain the spectrogram for each phone (here and in subsequent sections, the length of frame used is 128, and the increment is 64), and rescale all the spectrograms processed as described in 
Section~\ref{mfcccoeffs} 
(which may be of variable temporal length) to size $14 \times 15$. As described in Section~\ref{allVall}, we train $N(N-1) / 2$ classifiers. Ignoring the confusions mentioned above, we obtain a $0.595$ correct classification rate. The table below contains more details. For each phone, we list the labels that are assigned to it most often by the classifier, and the frequency with which the label is assigned. (i.e., for example, the label [w] is assigned to [r] phones $3 \%$ of time).

\begin{center}
\includegraphics[max width=\textwidth]{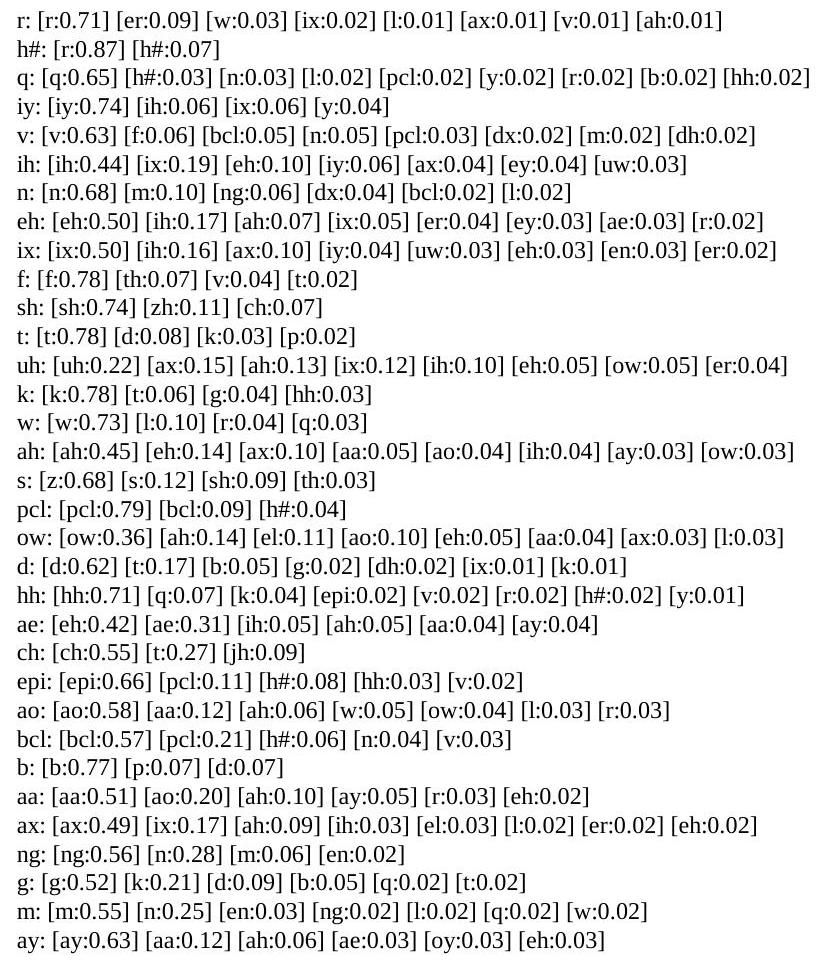}
\end{center}

\begin{center}
\includegraphics[max width=\textwidth]{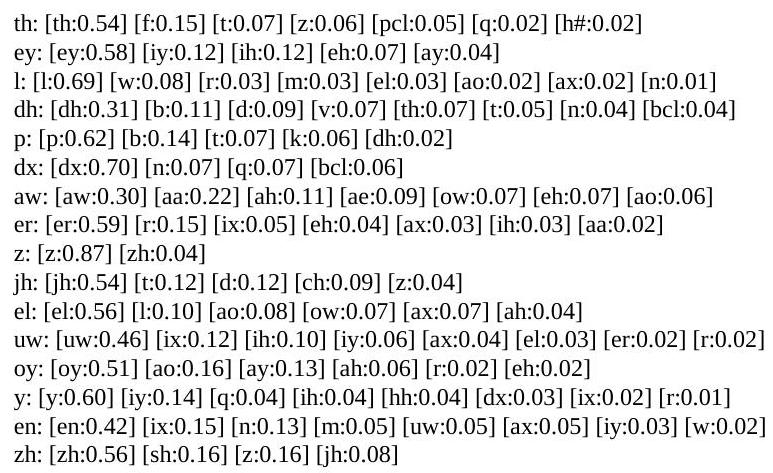}
\end{center}

We observe that confusion occurs mostly within phonetic categories (e.g., [t] is most often confused with [d], since they are both alveolar plosive, the former voiceless and the latter voiced, and hence have very similar spectrograms. On the other hand, [b] is never confused with $[\mathrm{s}])$.

We observe that the all-vs.-all classification scheme does not produce optimal results. The most obvious example is [h\#], which is almost always recognized as [r], even though the individual $[\mathrm{h} \#] /[\mathrm{r}]$ classifier outputs the correct result about $80 \%$ of the time, and [r] is almost never recognized as [h\#].

As discussed in Section~\ref{betterallVall}, we modified the all-vs.-all classifier to iteratively eliminate the phone that gets the fewest votes at each iteration. For N1 in the range $6 . .15$, we obtain $67 \%$ correct classification rates. For N1 $<5$, the performance deteriorates.

\subsection{Is there enough data? (Haar features, warped spectrograms)}
There is no way to tell whether performance would improve if more training data had been available. However, we can try varying the size of the training set and see whether the error is still decreasing when we are using all of the available training samples.

The following are the errors obtained (y-axis, in percent) when classifying t/d, s/sh, aa/ah, and m/n using classifiers trained with different sample sizes (x-axis, the number is the number of samples of each phone). 

\begin{figure}
aa/ah
 \begin{picture}(100,100)
  \put(0,-100){\includegraphics[max width=\textwidth]{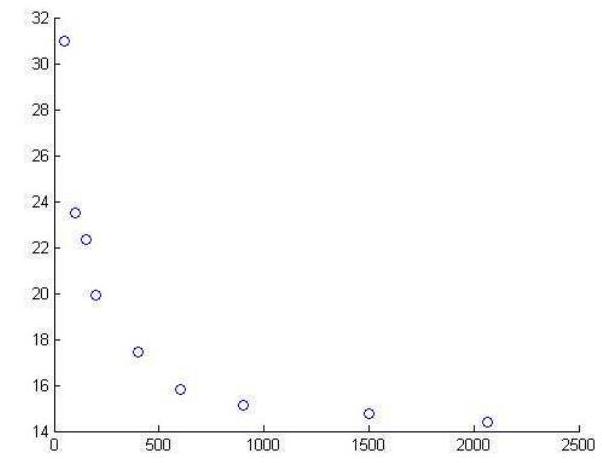}}
  \put(-20, 130){\% error}
  \put(250, -120){training set size}
\end{picture}
\end{figure}
\begin{verbatim}
  

 




\end{verbatim}

\newpage

\begin{figure}
  s/sh
   \begin{picture}(100,100)
    \put(0,-100){\includegraphics[max width=\textwidth]{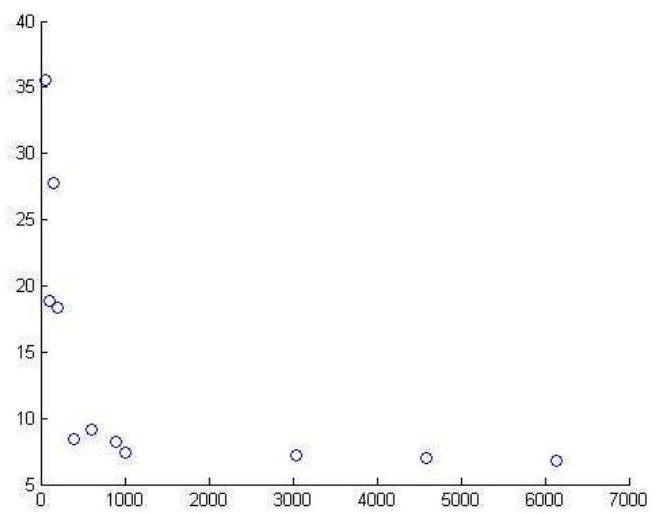}}
    \put(-20, 130){\% error}
    \put(250, -120){training set size}
  \end{picture}

\end{figure}
 \begin{verbatim}
  

 




 \end{verbatim}
 \newpage

\begin{figure}
  n/m 
   \begin{picture}(100,100)
    \put(0,-100){\includegraphics[max width=\textwidth]{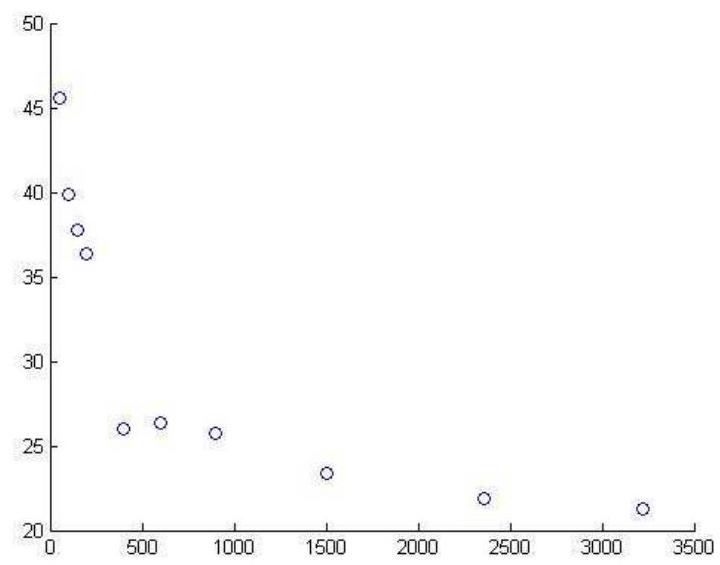}}
    \put(-20, 130){\% error}
     \put(250, -120){training set size}
  \end{picture}
\end{figure}
  
\begin{verbatim}
  

 



\end{verbatim}
\newpage
  
  \begin{figure}
  t/d
    \begin{picture}(100,100)
    \put(0,-100){\includegraphics[max width=\textwidth]{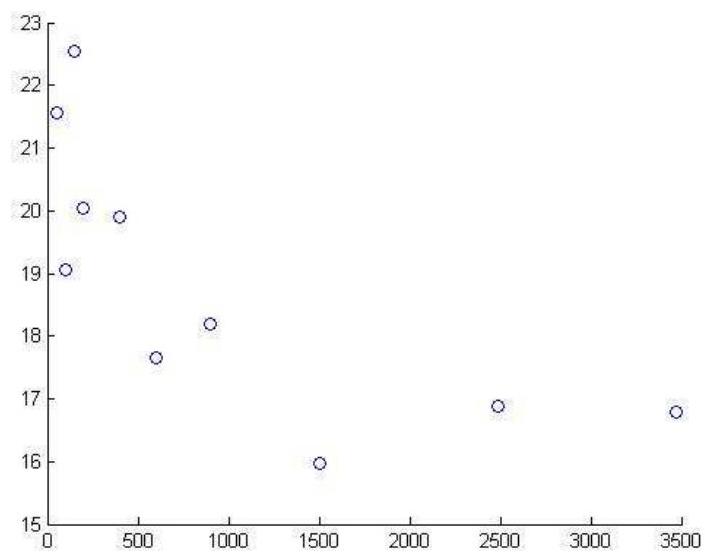}}
    \put(-20, 138){\% error}
    \put(250, -120){training set size}
   \end{picture}
 \end{figure}
  
 \begin{verbatim}
  

 



 \end{verbatim}
 \newpage















For aa/ah and m/n, it may be the case that having more data would cause the error to decrease further; the errors for t/d and s/sh appear to have stabilized before all the available training samples were used.

\subsection{Is there overfitting? (Haar features, warped spectrograms)}
The common claim about AdaBoost is that, unlike for most discriminative learning algorithms, making the classifier more complex (i.e., adding more classifiers to the linear combination) does not lead to the test error's increasing.

The following are the error rates when classifying using only the first $N$ selected classifiers (the x-axis) for several phones, with the test errors in blue and train errors in red.

\newpage

\begin{figure}[h]
  m/n
    \begin{picture}(100,100)
    \put(0,-100){\includegraphics[max width=\textwidth]{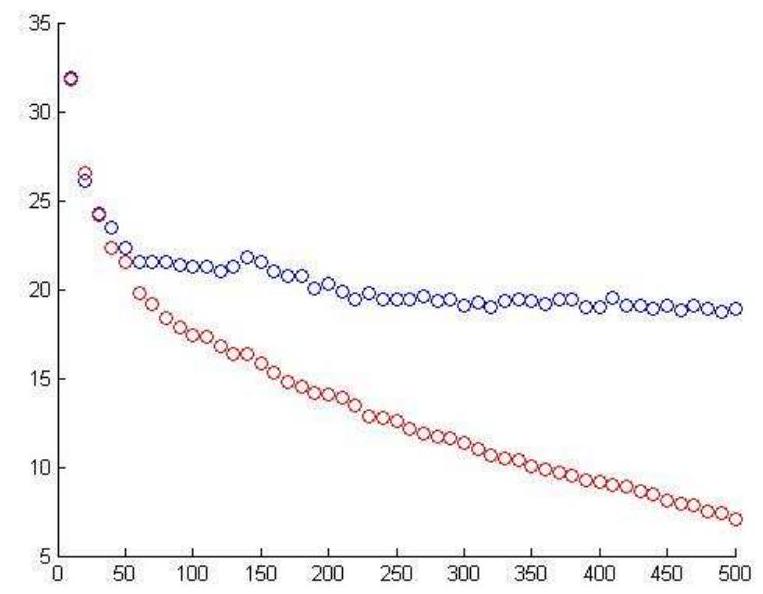}}
    \put(-80, 138){\% error (train and test)}
    \put(250, -120){number of classifiers}
   \end{picture}
 \end{figure}
  
 \begin{verbatim}
  

 



 \end{verbatim}
 \newpage

 \begin{figure}[h]
  aa/ah
    \begin{picture}(100,100)
    \put(0,-100){\includegraphics[max width=\textwidth]{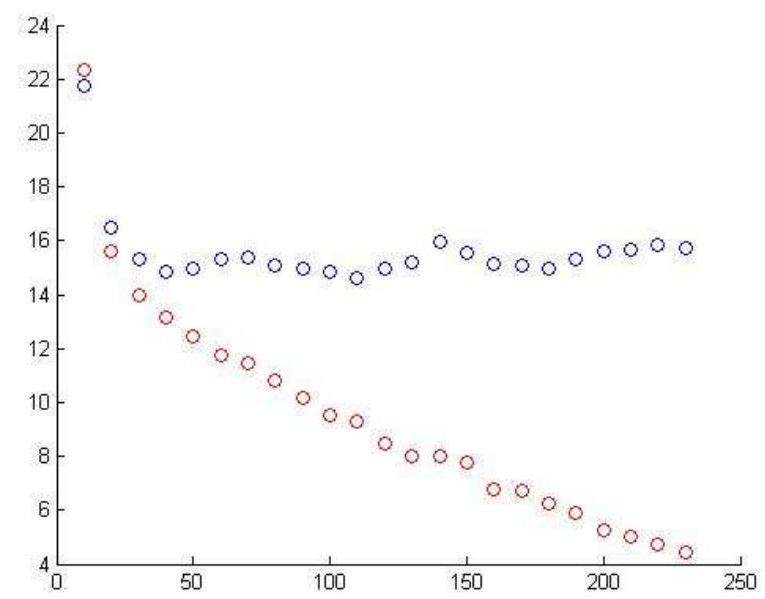}}
    \put(-80, 138){\% error(train and test)}
    \put(250, -120){number of classifiers}
   \end{picture}
 \end{figure}
  
 \begin{verbatim}
  

 



 \end{verbatim}
 \newpage

Although the training error decreases much faster than the test error, the test error does not substantially increase when we add more classifiers (note that the test error decreases very rapidly in the beginning.)

\subsection{Can we use context? (Haar features, warped spectrograms)}
Phones sometimes sound differently depending on the context due to coarticulation effects. It therefore makes sense to try to train on spectrograms that include not just the phone itself, but to also include the patches surrounding the phone. The resolution is not affected since we expand the width of the normalized sample proportionately to the new average length of the samples. The results are as follows:

$\mathrm{t} / \mathrm{d}$

\begin{center}
\begin{tabular}{|l|l|l|}
\hline
$\begin{array}{l}\text { Width of margins on either } \\ \text { side (seconds) }\end{array}$ & Training error (\%) & Test error (\%) \\
\hline
0.03 & 4.33 & 16.20 \\
\hline
0.04 & 4.93 & 16.11 \\
\hline
0.06 & 5.03 & 14.66 \\
\hline
0.08 & 4.76 & 15.12 \\
\hline
\end{tabular}
\end{center}

aa/ah:

\begin{center}
\begin{tabular}{|l|l|l|}
\hline
$\begin{array}{l}\text { Width of margins on either } \\ \text { side (seconds) }\end{array}$ & Training error (\%) & Test error (\%) \\
\hline
0.03 & 4.00 & 15.35 \\
\hline
0.04 & 4.97 & 13.42 \\
\hline
0.06 & 4.17 & 14.30 \\
\hline
0.08 & 4.26 & 15.12 \\
\hline
\end{tabular}
\end{center}

sh/zh

\begin{center}
\begin{tabular}{|l|l|l|}
\hline
$\begin{array}{l}\text { Width of margins on either } \\ \text { side (s) }\end{array}$ & Training error (\%) & Test error (\%) \\
\hline
0.03 & 7.07 & 23.38 \\
\hline
0.04 & 8.72 & 17.81 \\
\hline
0.06 & 8.72 & 20.35 \\
\hline
0.08 & 6.84 & 17.81 \\
\hline
\end{tabular}
\end{center}

$\mathrm{m} / \mathrm{n}$

\begin{center}
\begin{tabular}{|l|l|l|}
\hline
$\begin{array}{l}\text { Width of margins on either } \\ \text { side (s) }\end{array}$ & Training error (\%) & Test error (\%) \\
\hline
0.03 & 6.76 & 18.84 \\
\hline
0.04 & 6.07 & 18.56 \\
\hline
0.06 & 8.79 & 17.88 \\
\hline
0.08 & 6.03 & 18.73 \\
\hline
\end{tabular}
\end{center}

While we don't observe significant improvement, it seems that the additional data helps slightly with classification.

\subsection{Not using segmentation (as much) (Haar features) \label{nosegm}}
Phone segmentation is not generally available in speech data. We try to classify phones without knowing the exact boundaries of the phones. For each phone, we take the centre of the sound segment c, and then obtain the spectrogram for the segment [c$120 \mathrm{~ms}, \mathrm{c}+120 \mathrm{~ms}$. This spectrogram is then rescaled to $14 \mathrm{x} 15$ pixels. We proceed as in Section 6.1, and obtain $60.5 \%$ correct classification rate. This is significantly higher than the error rate obtained in Section 6.1. Segmentation of phones is very helpful for classification

\subsection{Not warping the spectrograms (as much) (Haar features)}
Here we provide evidence that not warping the spectrogram improves classification performance. We apply the $3^{\text {rd }}$ approach to dealing with the variable length of the the samples in Section ~\ref{varlength57} to selected pairs of phones, and compare the error rates to our baseline of warping all the spectrograms to size $14 \times 15$

\begin{center}
\begin{tabular}{|l|l|l|}
\hline
Pair & Baseline Test Error & Stacked Spectrogram Test Error \\
\hline
$\mathrm{m} / \mathrm{n}$ & $9.5 \%$ & $3.5 \%$ \\
\hline
$\mathrm{s} / \mathrm{z}$ & $19.9 \%$ & $19.7 \%$ \\
\hline
\end{tabular}
\end{center}

It seems that not warping the spectrograms to the same size significantly hurts performance for some pairs of phones.

We proposed one way to ameliorate this problem. What we propose seems roughly equivalent to building classifiers that only take as input samples of lengths that lie in a specific range.

\subsection{Using Histograms of Gradients}
We present some preliminary results for phone classification using Histograms of Gradients (HoG).

The methods for dealing with the variable length of samples that were used above do not work for HoG features: typically, we cannot do much better than a 50\% classification rate.

We use the fourth option from Section ~\ref{varlength57} for dealing with variable-length samples: for longer samples, the HoG feature we compute is actually the bin-wise average (or maximum) of the spread HoG features (we refer to these as avg-pooling and maxpooling respectively). An alternative to taking the average is taking the maximum. It is suggested by the results of Boureau et al.~\cite{boureau2010} that this may be the better alternative, but we find that it is not so on our dataset.

Possibly a better alternative is to average (or take the max of) the outputs of the SVM on the raw HoG features at the appropriate coordinates. However, it is not immediately clear how the SVMs should be trained, and we do not pursue this option here.

Since we compute the gradients for the HoG feature by simply computing the differences between adjacent pixels, we need to be careful to not have vertical striations in the spectrogram.

Similarly to what is described in Section ~\ref{varlength57}, we stack three spectrograms together, the first unmodified, the second with the rows convolved with the filter $[1, 2 ,1]$ and the third convolved with the filter $[1, 2, 5, 2, 1]$. This lets us preserve all the information in the spectrogram while also allowing AdaBoost to select features computed on a smoothed version of the spectrogram.

We obtain the following correct classification rates on the test set for two pairs of phones. We include the performance using the basic method reported in Section 6.6 for comparison.

\begin{center}
\begin{tabular}{|l|l|l|l|}
\hline
 & Max-pooling of HoG & Avg-pooling of HoG & Haar baseline \\
\hline
$\mathrm{m} / \mathrm{n}$ & $74 \%$ & $80 \%$ & $96.5 \%$ \\
\hline
aa/ae & $86 \%$ & $87 \%$ & $93.7 \%$ \\
\hline
\end{tabular}
\end{center}

 While the classification performance using HoG features is worse than what we obtain using Haar features, it is interesting that non-trivial HoG features are selected. For example, the eighth feature selected for $\mathrm{m} / \mathrm{n}$ classification using avg-pooling is of size $8 \times 4$, and the coefficients of the individual orientation bins are: $[-3.33, -3.42,  0.431,  .199,  1.30,  1.04, 2.26, -0.41, 0.12]$

  

Since the feature is larger than $2\times 2$, it is nontrivial to interpret. The fact that we have a set of large positive coefficients, a set of large negative coefficients, and a set of coefficients close to 0 means that this is not a simple edge detector.

Note that this is not necessarily a representative feature. Many of the features are small, and in most the positive and negative coefficients appear contiguously (mod 9).

\section{Conclusions and Future Work}
The correct classification rates we obtained are below those of Bouvrie et al.~\cite{bouvrie2008}. Unlike Bouvrie et al., we do not use the length of the sample as one of the features, so our results are not directly comparable.

We show that all-vs.-all classification can be improved by hierarchical voting

We have shown that SVM-classified Histrograms of Gradients (HoG) can be used as phonetic features, although Haar features seem superior to HoG features alone. It is possible that adding HoG features to the pool of Haar features would improve classification. We have shown that HoG features which are not interpretable as simple edge detectors are useful for classifying phones.

We have left trying replicating Haar features along the length of the spectrogram as a way of achieving 1-1 correspondence between features on spectrograms of different lengths (method 4 in Section ~\ref{varlength57}) for future work. Since this was helpful with HoG features, it is likely that doing so would lead to improved performance. Various image feature sets were introduced by the object detection community (e.g., Sabzmeydani et al~\cite{sabzmeydani2007} and Tuzel et al.~\cite{tuzel2007}.) These might prove useful for classifying phones.

For the phone classification task, there is no reason to expect that there is one best way to build a classifier that fits all $\mathrm{N}(\mathrm{N}-1) / 2$ pairs of phones. Optimizing a real system may involve hand-crafting each if the $\mathrm{N}(\mathrm{N}-1) / 2$ classifiers.

After good phone classification results are achieved on segmented data, the next step is try to use the same features to classify unsegmented speech data like we did in Section~\ref{nosegm}.

\bibliography{msc}
\bibliographystyle{plain}

\end{document}